\documentclass[letterpaper]{article}
\usepackage{aaai}
\usepackage{times}
\usepackage{helvet}
\usepackage{courier}
\usepackage{graphicx}

\begin{document}

\title{Formal Fields: A Framework to Automate Code Generation Across Domains}
\author{Jacques Basald{\'u}a\\
Lead author of the Jazz platform\\
Senior Data Scientist @ BBVA Data \& Analytics}

\maketitle

\begin{abstract}
\begin{quote}
Code generation, defined as automatically writing a piece of code to solve a given problem for which an evaluation function exists,
is a classic hard AI problem. Its general form, writing code using a general language used by human programmers from scratch is thought
to be impractical. Adding constraints to the code grammar, implementing domain specific concepts as primitives and providing examples
for the algorithm to learn, makes it practical.

Formal fields is a framework to do code generation across domains using the same algorithms and language structure. Its ultimate goal
is not just solving different narrow problems, but providing necessary abstractions to integrate many working solutions as a single
lifelong reasoning system. It provides a common grammar to define: a domain language, a problem and its evaluation. The framework
learns from examples of code snippets about the structure of the domain language and searches completely new code snippets to solve
unseen problems in the same field.

Formal fields abstract the search algorithm away from the problem. The search algorithm is taken from existing reinforcement learning
algorithms. In our implementation it is an apropos Monte-Carlo Tree Search (MCTS).

We have implemented formal fields as a fully documented open source project applied to the Abstract Reasoning Challenge (ARC). The
implementation found code snippets solving twenty two previously unsolved ARC problems.
\end{quote}
\end{abstract}

\section{1. Introduction}

In this introduction we briefly review some relevant work in both code automation and machine reasoning. In section 2 we introduce
formal fields as a set of necessary elements to do code execution across domains. In section 3 we have a closer look at the language.
In section 4 we study the efficiency of the code searching algorithm and point out other ideas not yet implemented. In section 5 we
summarize results of the implementation to the ARC challenge. Full results are available on github \cite{basaldua2020arc}. Finally,
we discuss how evaluating code in the context of formal fields leads to a pragmatic definition of understanding and how important the
distinction between not knowing and not understanding becomes if we intend to build multi-domain lifelong learning systems.

\subsection{1.1 Automated code generation}

Automated code generation, under different names, is one of the oldest problems in AI. In the 70s \cite{manna1971toward} suggest
theorem-proving to define specifications for automated programming and induction to construct programs with loops and recursion. You can
find more historical examples in the introduction of \cite{devlin2017robustfill}.

With the recent advances of Deep Learning (DL), the field, mostly under the name neural program synthesis, has experienced noteworthy
progress. Key ideas start with neural Turing machines \cite{graves2014neural}, which extend the capabilities of neural networks by coupling
them to external memory resources which they can interact with by attentional processes. \cite{zaremba2015reinforcement} add reinforcement
learning to the idea and \cite{devlin2017robustfill} double the attention mechanism to adapt to noisy conditions.

It is worth noting that in code generation, DL is not as dominant as in other AI fields. If we clearly separate the searching background
and compare different algorithms \cite{gaunt2016terpret} including linear programming and discrete satisfiability solving, we see that
"classical" approaches still play a major role. Best approaches are probably hybrid. In our implementation we do not use DL, but identify
specific targets for deep reinforcement learning in future implementations.

A survey paper on the field \cite{kant2018recent} concludes: "While it is still far from being solved or even competitive with most
existing methods, neural program synthesis is a rapidly growing discipline which holds great promise if completely realized."

Applying Reinforcement Learning (RL) to a program execution framework using NN is described in \cite{graves2016hybrid} and there is also
previous work on learning to code from examples in \cite{zaremba2016learning}. Using MCTS as a general optimizing algorithm to do AutoML
is described in AutoML-Zero \cite{real2020automl} and MOSAIC \cite{rakotoarison2019automated}.

For completeness sake, we mention approaches focusing on computer languages used by humans. Deep coder \cite{balog2016deepcoder} predicts
properties of the program by fitting outputs from inputs end to end with a neural network. Stratified synthesis \cite{heule2016stratified}
synthesizes the semantics of instructions whose semantics are unknown by bootstrapping from a set of instructions whose semantics are known.
TransCoder \cite{lachaux2020unsupervised} is a transpiler using unsupervised machine translation across high level languages trained on
large data from source code repositories. Using a constraint solver \cite{pu2017selecting} fit Python programs to match input-output
examples.

In this article we use the term Automated Code Generation (ACG), since the neural part is not essential, and narrow it down to the
case where:
\begin{enumerate}
  \item The code is generated as a solution to a given problem. The solution returned by the code can be evaluated by some function.
  \item The code can be explicitly described in source form. It is a sequence of instructions rather than a set of arguments or weights
  used by some other program.
  \item Once found, the code runs independently and its resource use is similar to an efficient human written program.
\end{enumerate}

\subsection{1.2 Machine reasoning and lifelong learning}

In the discussion we advocate how formal fields are instrumental to build complex lifelong reasoning systems. This builds on top of some
ideas we briefly introduce here.

Semantic computation can be traced back to the origins of AI with Quillian's semantic memory\cite{minsky1968semantic}. Later semantic
networks \cite{sowa1987semantic} lead naturally to the need of a description logic \cite{baader2003description}. A classic approach to
computer semantics has been storing information as semantic triples, e.g., \cite{miller1998introduction} and using some formal reasoning
system to do inference with them. As the database grows, it inevitably becomes brittle, among other things because of its inability to
manage uncertainty. There is also a deeper problem, the Symbol Grounding Problem (SGP). In the words of \cite{harnad1990symbol}, it is
the problem of making ”the semantic interpretation of a formal symbol intrinsic to the system, rather than just parasitic on the meanings
in our heads.” Despite claims about the SGP being solved, after reviewing fifteen years of SGP research, \cite{taddeo2005solving} concluded
"All the strategies are semantically committed and hence none of them provides a valid solution to it."

Shortcomings of current approaches to DL have been pointed out recently, e.g., \cite{marcus2018deep}. We precise "current approaches"
because it could be argued that many limitations are not inherent to DL but related to how we use (and not re-use) learning. In an
influential work \cite{bottou2014machine} suggests a way to keep the best of both approaches: "Instead of trying to bridge the gap between
machine learning systems and sophisticated 'all-purpose' inference mechanisms, we can instead algebraically enrich the set of manipulations
applicable to training systems, and build reasoning capabilities from the ground up." In this direction, some efforts include
\cite{garnelo2016towards} and \cite{hudson2019learning} in the field of visual reasoning. The necessity to scale up to complex systems by
using just one model is also identified in "One Model To Learn Them All" \cite{kaiser2017one}, a model architecture containing
convolutional layers and an attention mechanism for multiple domains.

Formal fields are extremely efficient to run and hard to search. Therefore, reusing learned knowledge is key to building complex
formal field based systems. This introduces the need for lifelong learning \cite{thrun1998lifelong}. In the same book chapter we find the
following definitions:

\begin{itemize}
  \item \emph{Learning}: Given a task, training experience and a performance measure, a computer program is said to learn if its
  performance at the task improves with experience.
  \item \emph{Lifelong Machine Learning} (LML): LML, considers systems that can learn many tasks over a lifetime from one or more domains.
  They efficiently and effectively retain the knowledge they have learned and use that knowledge to more efficiently and effectively learn
  new tasks.
\end{itemize}

Recent work \cite{silver2013lifelong} advocates for "more seriously considering the nature of systems that are capable of learning over
a lifetime."

A phenomenon (not only) related with lifelong learning is catastrophic forgetting. Originally known as catastrophic interference, it is
described as "New learning may interfere catastrophically with old learning when networks are trained
sequentially."\cite{mccloskey1989catastrophic}. The problem is addressed with different approaches, e.g., \cite{d2019episodic}.

\section{2. The formal fields framework}

The formal fields framework is a compilation of all the necessary elements to do ACG, both running code and searching programs that
solve problems.

\subsection{2.1 Prerequisites}

We discuss language details later, for the definition of formal fields, language types, primitives and structure are introduced here. Table
\ref{table:prelim} summarizes the prerequisites and introduces the notation.

\subsubsection{2.1.1 Types}

All arguments and returned values are strictly typed. All types except tuples are tensors. Tuples are ordered structures combining any
types, including other tuples. Types support inheritance. Inheritance is also used implicitly: e.g., Tensors with a shape are inherited
from generic tensors, numerical conversions are automatic from integer to floating point and from a smaller precision to a bigger precision.
Inherited types can be used as arguments for all their ancestor types, but not the other way round. I.e., the behavior is covariant in the
return types and contravariant in the arguments.

\subsubsection{2.1.2 Primitives and FSL}

A \emph{primitive} is a function. Primitives are strictly typed. The set of primitives in a field is called the
\emph{Field Specific Language} (FSL).

\subsubsection{2.1.3 Kinds}

A \emph{kind} is a data type that serves a purpose across the whole system. It can also have its own primitives. If it does, they will be
included in the FSL. It has some resemblance with a class in computer languages (combining code and data), but, unlike objects of a
class, instances of a kind are just data and have no methods.

For example: A wave is a kind. Speech is a wave. Speech to text, text to speech, speech to speech, are examples of fields.

\subsubsection{2.1.4 Domain (X) and range (Y)}

A field has two special kinds: $X$ and $Y$ also referred to as \emph{domain} and \emph{range}. There is no limit on how complex a kind can
be or how many types the primitives in the FSL use. These two kinds are special because they define both ends of a code snippet. A code
snippet $p$ is a function $p(x \in X) \to Y$.

\subsection{2.2 Code structure}

In the formal fields framework \emph{code} is a sequence of instructions that runs once. Conditional branching and iteration is done as
discussed in section 3. The mechanism used to pass arguments to functions is a \emph{stack}. Any execution tree can be written as a
sequence. In higher level languages compilers translate source code into sequences of instructions. Both are functionally equivalent.
Writing sequences directly is less human friendly, but still human understandable and removes overhead. A language \emph{kernel} includes
primitives for stack manipulation that are available in the FSL of any field: E.g., pushing constants, splitting tuples, swapping top
elements, \dots

\subsubsection{2.2.1 Primary structure: opcodes}

An \emph{opcode} is an executable code instruction. An opcode can be a primitive, which is a function, or a constant, which can be viewed
as a function without arguments defined in the kernel.

\subsubsection{2.2.2 Cores}

A function that executes code is called a \emph{core}. Code, in general, is any sequence of opcodes. There is a kernel function
$\hat{c}(s, o\star, L) \to t \in T$ where $s$ is a state of the stack, $o\star$ is a sequence of opcodes and $L$ is a FSL. The initial state
of the stack $s$ contains an element $x \in X$. This function can return any type in $T$, the set of types used in $L$.

The core checks types and errors in function calls and immediately stops on error. An error is just a kernel type and is in $T$.

In a field, this function becomes $c(s, o\star) \to Y\star$ where $Y\star$ is the set of sequences of $Y$. Note that the language is
part of the field and not an argument anymore. More importantly, we are only interested in the opcodes that return $y \in Y$ and we are
interested in all of them, not just the last one. Possible constants of type $Y$ in the code are not included as results.

\subsubsection{2.2.3 Tertiary structure: code snippets}

A \emph{code snippet} is a sequence of opcodes $p \in O\star$ such that $\hat{c}(x, p, L) \in Y$ This implies that, at least for that given
$x \in X$ it produces an error free result. The snippet is what we would less precisely call a program. ACG is about finding code
snippets. We also define $P$ as the space of all snippets.

\subsubsection{2.2.4 Secondary structure: code items}

A \emph{code item} is a minimal snippet. I.e., a snippet that only has one opcode, the last one, returning an $y \in Y$. Code snippets are
built as sequences of items rather than one opcode at a time. Unlike code snippets which run end-to-end from $X$ to $Y$, items run from
some initial stack state $s$ to $Y$. This state $s$ is, initially, some $x \in X$ and later becomes the stack state left by items running
previously.

\begin{table}[h!]
\centering
\begin{tabular}{|c c|}
\hline
\multicolumn{2}{|c|}{Sets} \\
\hline
$O$ & Set of opcodes \\
$O\star$ & Set of sequences of $o \in O$ \\
$X$ & Domain (a kind) \\
$Y$ & Range (a kind) \\
$Y\star$ & Set of sequences of $y \in Y$ \\
$L$ & Set of primitives (the FSL) \\
$T$ & Set of types in L \\
$P$ & Set of all code snippets \\
$S$ & Set of stack states \\
\hline
\multicolumn{2}{|c|}{Core functions} \\
\hline
$s \in S$ & A stack state \\
$o\star \in O\star$ & A sequence of opcodes \\
$\hat{c}(s, o\star, L) \to t \in T$ & Core in the framework \\
$c(s, o\star) \to Y\star$ & Core in a field \\
\hline
\multicolumn{2}{|c|}{Code structure} \\
\hline
$x \in X$ & Element in domain \\
$q \in P \; \slash \; |c(s, q)| = 1$ & Code item \\
$p \in O\star \; \slash \; \hat{c}(x, p, L) \in Y$ & Code snippet \\
\hline
\end{tabular}
\caption{List of prerequisites}
\label{table:prelim}
\end{table}

\subsection{2.3 Formal field}

A tuple $(X, Y, L)$ where $X, Y$ are two kinds and $L$ is a FSL is called a \emph{formal field}. The field has other things provided
by the framework, e.g., a core and a kernel language that are shared by all the fields in the framework. The field provides the complete
functionality to run code across two kinds. The implementation also provides some utilities to compile and decompile code to a human
readable format, save and restore state to compute trees of snippets reusing previous runs, etc.

\subsection{2.4 Formal relation}

\emph{Formal relations} implement supervised learning using formal fields. Here we systematize the necessary parts, in section 4 we
describe how search works in our implementation.

\subsubsection{2.4.1 Evaluation function}

A function $e(\hat{y}, y) \to {\rm I\!R}$ used to compare two elements in the range: the computed $\hat{y}$ and some ground truth $y$ is
called an \emph{evaluation function}. Note that this function operates element-wise. It can be aggregated by some other function over
a complete dataset. Also note that there is no restriction on how complex an element is, an element can be a pixel, an image or a batch
of images.

\subsubsection{2.4.2 Codebase}

A \emph{codebase} $B$ is a set of $N$ snippets $p_i \in P$ and examples $x_i \in X$ such that
$\forall i \in {1,..N} \; p_i(x_i) = y_i = \widetilde{y}(x_i)$ all snippets run on their example $x_i$ and produce the same result
a hypothetical ground truth function $\widetilde{y}(x_i)$ would.

The codebase is used to create an initial set of code items by splitting the snippets and applying mutations to the resulting items.
The codebase is also used to establish prior values for the original and mutated items.

\subsubsection{2.4.3 Prior function}

A function $u(q, B) \to {\rm I\!R}$ where $q$ is a code item and $B$ is a codebase is called a \emph{prior function}. During search,
each node defines some code item appended to a code snippet (the shortest path from root to its parent). Prior functions assign prior
values to nodes. These prior values, combined with rewards, determine the order in which nodes are visited.

\subsubsection{2.4.4 Value function}

A function $v(x, p, \widetilde{y}(x)) \to M$ where $x \in X$ is an element in the domain, $p \in P$ is a code snippet,
$\widetilde{y}(x) \in Y$ is the ground truth value corresponding to $x$, and $M = {\rm I\!R}^k$ is a \emph{value space} of dimension $k$,
is called a \emph{value function}. Value spaces are multivariate because univariate representation cannot express the complexity of
intermediate values (see 4.1). Note that the value function not only evaluates the final $\hat{y} = p(x)$, but all the $y\star = c(x, p)$
returned by each item in the snippet. The value $v \in M$ of a node is converted later into a reward by the reward function.

A value function can be handcrafted, using such things as absolute evaluation: $e(p(x), y)$, improvement by the last item
$e(\hat{y}_t, y) - e(\hat{y}_{t-1}, y)$, \dots. An element $x$ can include more than one examples, resulting in more computed values over all
the examples (worst, mean, best, \dots) for each of the previous, etc. Besides handcrafting, a deep neural network approach is suggested
in section 4.

\subsubsection{2.4.5 Reward function}

A function $w(v) \to {\rm I\!R}$ where $v = v(x, p, y)$ is a value $m \in M$ is called a \emph{reward function}. The reward is used to
guide search. After expanding a leaf, it propagates backwards from the last visited leaf up to the root node. The reward function is
learned from a dataset built from the codebase (positive rewards) adding random mutations of snippets and snippets running on wrong domain
elements as examples of negative rewards. In our implementation, the learning algorithm is Xgboost \cite{chen2016xgboost}.

\subsubsection{2.4.6 Formal relation}

A tuple $(F, B, e, u, v, w)$ where $F$ ia a formal field, $B$ is a codebase in that field, $e, u, v, w$ are: evaluation, prior, value
and reward functions is called a \emph{formal relation}. A relation is all you need to do ACG. All other things: cores to run the snippets
and a search algorithm are implemented in the framework for all relations in a multi-field system. A field can be used in more than one
relation. Table \ref{table:relation} summarizes the complete framework.

\begin{table}[h!]
\centering
\begin{tabular}{|c c|}
\hline
\multicolumn{2}{|c|}{Formal field} \\
\hline
F & $(X, Y, L)$ \\
\hline
\multicolumn{2}{|c|}{Codebase} \\
\hline
$B$ & $(p_i \in P$, $x_i \in X), i \in {1,..N}$ \\
\hline
\multicolumn{2}{|c|}{Other} \\
\hline
$\hat{y} \in Y$ & Predicted value \\
$y \in Y$ & Ground truth value \\
$k$ & Dimension of the value space \\
$M$ & Value space ${\rm I\!R}^k$ \\
$m \in M$ & Value \\
\hline
\multicolumn{2}{|c|}{Functions} \\
\hline
$e(\hat{y}, y) \to {\rm I\!R}$ & Evaluation function \\
$u(q, B) \to {\rm I\!R}$ & Prior function \\
$v(x, p, y) \to M$ & Value function \\
$w(m) \to {\rm I\!R}$ & Reward function \\
\hline
\multicolumn{2}{|c|}{Formal relation} \\
\hline
(F, B, e, u, v, w) & Field, codebase and functions \\
\hline
\end{tabular}
\caption{The formal field framework}
\label{table:relation}
\end{table}

\section{3. Bebop, a single-pass ACG language}

The field of designing formal languages intended for symbolic computation and creating programs by other programs dates back to the origin
of Lisp \cite{mccarthy1978history}. Unlike search (of programs or whatever) where a sound mathematical theory can be referred to as an
optimal (most of the times impractical) gold standard, designing a language is based on simplifying and extending previous designs.
In the case of languages for ACG, besides Lisp there are not many examples. Bebop is the implementation of the ideas described here.
Bebop has a fully functional proof of concept implementation in Python \cite{basaldua2020arc} and an in-progress industrial level C++
implementation \cite{basaldua2017jazz}.

\subsection{3.1 Language design considerations}

The language grammar implemented in the framework is \emph{minimalistic} and \emph{predictable resource-wise}. This comes at a price as we
discuss here.

\subsubsection{3.1.1 Curbing hypothesis space}

When exploring our \emph{hypothesis space} $P$, the most important hurdle comes from Church \cite{church1936unsolvable} and
Turing \cite{turing1937computable}. The \emph{halting problem}, at the time known as Entscheidungsproblem. As Turing immediately
realized: "In a recent paper Alonzo Church has introduced an idea of \emph{effective calculability}, which is equivalent to my
\emph{computability}, but is very differently defined. Church also reaches similar conclusions about the Entscheidungsproblem.
The proof of equivalence between \emph{computability} and \emph{effective calculability} is outlined in an appendix to the
present paper."

Oddly, the key idea that gives sense to AI, the idea that all computation systems, including biological, are equivalent also introduces
the biggest hurdle: We have no way to know what a program will return other than running it. It may take infinite or practically
infinite time to run and that running time is unpredictable too.

We could limit the number of instructions a program can execute before we arbitrarily declare it failed, or similar ideas. But, we
considered a simpler although extreme measure: \emph{removing conditional execution completely}.

\subsubsection{3.1.2 How conditionals are implemented}

A good indicator that the idea is feasible are biological systems. Biological systems are built on top of code which is a sequence
of codons (groups of three base pairs) executed in a single pass. The way nature does conditionals is using \emph{inhibition},
\emph{epigenetics} and other mechanisms. This is equivalent to setting weights to zero (inhibition) or connecting inputs of artificial
neurons to memory cells representing some state (epigenetics). It can be done using just arithmetic without conditional jumps.

Another way to do conditionals is exploiting the richness of the type system. On the simplest level, a primitive can do things like
detecting all the red elements in a picture returning them as a tuple, another primitive can take this tuple and filter only those
vertically symmetric, etc. The language is just defining a pipeline of primitives that iterate internally.

A more complex example, like a shape moving over an image and stopping when some condition happens, can be represented by a primitive
that takes a tuple of (shape, position, movement) and returns a tuple of the same type. Initially, movement is a vector indicating
a direction. After the condition is satisfied, movement becomes (0, 0) and all subsequent executions do nothing.

Finally, iteration is an important use of conditional branching. Having no explicit looping instructions is not a functional limitation,
unrolling loops is a common program optimization technique. Formal fields search code snippets by adding code items to existing snippets.
Since there is no limit on how complex an item can be, there is no limit on how complex a loop can be. Therefore, when the evaluation sees
some advantage in adding the same item again and again rather than doing something else, a snippet will be doing iteration, possibly
combined with inhibition. Many algorithms will not show this gradually improving behavior that favors finding them. This evolutionary
approach to coding finds solutions that are different from what humans would code, a usual behavior of machine learning algorithms.

\subsubsection{3.1.3 Failing fast}

As mentioned, snippets have to run to be evaluated. There are two major incentives in running fast: 1. Snippets are the final product used
for building lifelong learning systems and 2. Search should be able to run billions of snippet candidates on reasonable experimental
settings.

Error is a normal return value for code and should be forced as early as possible. The kernel implements a Halt and Catch Fire (HCF)
instruction that immediately breaks returning an error type. Primitives should use it as much as possible. Errors are raised by type
checking and should also be raised in primitives by unexpected argument values, including just something becoming too big. E.g., There may
be primitives repeating something n times. These primitives running on their own results can easily produce something gigantic, breaking it
as soon as possible is essential.

\subsubsection{3.1.4 Expressiveness vs. practicability tradeoff}

By removing conditional execution we have completely avoided the halting problem and made the hypothesis space "just" a combinatorial
problem, like most machine learning problems. Assuming we have: structure, a fitting mechanism and the space contains enough "good
solutions", the method is practical.

More flexible languages would be more elegant human-wise and achieve more with less code, Lisp being the best example, but besides
textbook implementations of trivial programs extending themselves, we do not know how to code in these languages automatically. There
are examples of translation across languages (formal to formal) or fitting parametrized programs to given outputs, but that does not
match the definition of ACG (informal to formal, i.e., intelligent).

We call this the "expressiveness vs. practicability tradeoff": We decided to go \emph{practical first} and later maximize expressiveness
without compromising practicability.

Minsky's influential essay on language design \cite{minsky1967programming} already stated how expressiveness can build on top of rigid
grammar, just here "programmer" is a machine. "A rigid grammar need not make for precision in describing processes. The programmer must
be very precise in following the computer grammar, but the content he wants to be expressed remains free."

\subsubsection{3.1.5 Primitives, combination and abstraction}

A classic textbook on formal languages \cite{abelson1996structure} arranges design considerations in:

\begin{enumerate}
  \item \emph{Primitive expressions}: which represent the simplest entities the language is concerned with.
  \item \emph{Means of combination}: by which compound elements are built from simpler ones.
  \item \emph{Means of abstraction}: by which compound elements can be named and manipulated as units.
\end{enumerate}

Bebop's minimalistic design in those terms is: \emph{Primitive expressions} are at the core built on top of type inheritance and tuples.
The \emph{means of combination} is a sequence of opcodes that runs once. This is functionally equivalent to a no-conditional execution
tree. Finally, the \emph{means of abstraction} are: 1. the idea of code items and 2. primitives written in Bebop, supported only in the
Jazz implementation.

\subsection{3.2 More on program structure}

We already introduced primary structure (opcodes), secondary (items) and tertiary (snippets) formally in section 2. Here we describe
some ideas that are key to create code items using the snippets in the codebase.

The first step is splitting the snippets. Breaking the sequences to make them finish at every non constant opcode returning an $y \in Y$.
Then, building an "item base" with precomputed priors.

\subsubsection{3.2.1 Alleles and constant fitting}

Two code items that are identical in their primitives but differ in the values of constants are known as \emph{alleles}. When an item
contains constants, that opens the question of constant fitting. The current implementation has two approaches to it: 1. Just copying
successful alleles as they are, possibly more than one letting them "compete" in the search. 2. Only in case the evaluation function can
identify the ground truth $y$ to be reachable from the last computed element $y_t$, the evaluation function can modify the item. E.g., if
the solution to a problem is almost correct, just what is blue should be yellow, the evaluation function can append a "and recolor blue to
yellow" suffix, "blue" and "yellow" being some constant arguments passed to a "recolor" primitive. Otherwise, once a code item is inserted
in the search tree it is not modified. This is an opportunity for future improvement.

\subsubsection{3.2.2 Form and isomorphism}

The sequence of argument types and return types in a code item is known as its \emph{form}. Two code items sharing the same form are
known as \emph{isomorphisms}.

There is a necessity to explore items that are not in the codebase. This is done by creating new items changing functions of existing
items one at the time by other functions in the FSL that maintain the form. Other mutations include insertion and deletion of opcodes.
These "never seen before" items have their priors assigned based on similarity to successful items. This may look very limited (one
mutation only), but it has to be seen as a lifelong learning process. New items are tried and hopefully included in successful snippets
and end up in the codebase. In the next generation we see mutations of successful mutations. Exploring possibly malformed items
exploits the robustness of the framework: Everything in the tree runs from root, what doesn't, doesn't make it into the tree.

\section{4. Generating code by search}

We build code snippets by searching a tree starting at an empty root node. Every node represents appending a code item, and we have a
\emph{formal relation} $R$ which implies:

\begin{enumerate}
  \item We have a \emph{formal field}, $X \to Y$ with a FSL and can run code on it.
  \item We have a set of \emph{code snippets} $B$ from which we built a set of \emph{code items} $I$ as described in 3.2 with prior values
defined over all items by a prior function $u(q, B) \to {\rm I\!R}$.
  \item We have a value function $v(x, p, y) \to M$ and a reward function $w(m) \to {\rm I\!R}$ to guide the search.
  \item We have an evaluation function $e(\hat{y}, y) \to {\rm I\!R}$ to compare the output of a snippet $p(x)$ with some ground truth $y$.
\end{enumerate}

\subsection{4.1 Efficiency of the search}

After tackling the halting problem, we have to reckon with two more hurdles in order to understand ACG:

\begin{enumerate}
  \item Even if running an item is a deterministic function, when we abstract the field away from the search, at the search level we have
no understanding of what the program is doing. We cannot expect the program to reach a solution by gradually approaching the expected $y$
like an optimization algorithm would do on a regular problem. We have to model that uncertainty (the search's ignorance on what the
program is doing) as a Markov Decision Problem (MDP). Basically, accepting that what we evaluate at some stage $t$, the $y_t* \in Y*$, is
not the real state of the problem, but just a proxy. The real state would be described by some \emph{strong metric} like the length of the
optimal snippet that reaches $y$ from $y_t$. Such \emph{strong metric} is, of course, not computable.
  \item The only framework known to determine the value of our decision problem (what is the best item to append at each node) is based
on backward induction. We can, at best, apply it to already found snippets, but it is not computable during search. Gathering knowledge
about the "regularity" of the evaluation along the items of a (successful vs. unsuccessful) snippet is what we do when we learn the reward
function using the codebase. This explains why learned reward functions work better than handcrafted ones, it is not about idealizations
on how to reach a goal, it is about learned knowledge from the snippets that do vs. those that don't.
\end{enumerate}

Also, we don't even know if a snippet such that $p(x) = y$ can be built with the DSL and the codebase at all, but we presume that we
could always extend the language and the codebase to make that possible.

\subsubsection{4.1.1 UCB as an approximation to Bellman value}

The analysis of the value of each decision at a node (state) in terms of the immediate reward and the discounted  values of its successor
states is Bellman's equation \cite{bellman1957dynamic} formulated for a MDP \cite{bellman1957markovian}.

As mentioned, it is not computable in our case. The reason we refer to Bellman's equation is because what we use instead, Monte-Carlo
methods, should be viewed as an approximation to it.

As stated in \cite{sutton1998reinforcement}: "We would like to apply the approach of stochastic gradient descent (SGD) of the Bellman
error, in which updates are made that in expectation are equal to the negative gradient of an objective function. These methods always
go downhill (in expectation) in the objective and  because of this are typically stable with excellent convergence properties. Among the
algorithms investigated so far in this book, only the Monte Carlo methods are true SGD methods."

When traversing the tree forward, the selection of the next node to visit uses the UCB formula \cite{kocsis2006bandit} which is an evidence
based balance between exploration and exploitation using a confidence interval of the binomial proportion.

At every node $s\star$, we select his child $s_i$ that maximizes:
\begin{equation}
u(h + \log((n\star + f)/f))\sqrt{n\star}/(n + 1) + g(r/n)
\end{equation}

where $u$ is the prior value of the child, $n$ is the number of visits of the child, $r$ is the reward of the child, $n\star$ is the number
of visits of the parent and $f, g, h$ are constants.

Under the following assumptions:

\begin{enumerate}
  \item The reward is a true reward, e.g., a game won or lost rather than a value assigned by some learned reward function.
  \item Or, at least, in expectation the rewards obtained from random walks starting at a node are proportional to the probability of
the node being part of a successful snippet. In other words, the stochastic process, when adjusted to be a probability is an unbiased
estimator of success.
\end{enumerate}

The UCB formula (1) is a true approximation of the Bellman value. And that approximation will have a mean absolute error proportional to
$1/\sqrt{n}$ where $n$ is the number of visits through the node.

\subsubsection{4.1.2 The need for regularization}

Another central question in supervised learning is: We want our model (snippet) to perform well on unseen data. Techniques like
cross-validation are an external layer to what we are discussing in this paper and should be used in combination with the formal fields
framework. Here, we have a closer look at what overfitting is and describe a simple regularization technique used in our implementation.

By aggregating the evaluation $e(\hat{y}, y) \to {\rm I\!R}$ over a set of predictions $\{\hat{y}_i\} \in Y$ comparing them element-wise
with their ground truth values $\widetilde{y}_i \in Y$, we define a \emph{loss function} $l(y) \to {\rm I\!R}$.

For three snippets, $p(x)$ the best snippet found, $\widetilde{p}(x)$ the best snippet in $P$ and our hypothetical ground truth function
$\widetilde{y}(x)$, we can decompose the loss of $p(x)$ as: $$l(p(x)) = l(\widetilde{y}(x)) + \Lambda + \lambda$$ where
$\Lambda = l(p(x)) - l(\widetilde{p}(x))$ is difference between the loss of the snippet and the loss of the best snippet existing in P and
$\lambda = l(\widetilde{p}(x)) - l(\widetilde{y}(x))$ is the difference between the loss of best snippet in P that of the ground truth
function.

Note that $\lambda$ is the term representing the shortcomings of the FSL and the codebase, i.e., the \emph{underfitting}, the lack of
expressiveness and $\Lambda$ represents the shortcomings of our search, the \emph{overfitting}. Overfitting is a function growing with the
degree of freedom in the snippet and decreasing with the degree of freedom in the range. The first idea to reduce overfitting is shortening
the snippet.

The constants in the UCB equation 1 allow for adjustment between exploration and exploitation. Increasing exploration (going wide) tends
to make shorter snippets, but the optimal balance may be compromised if exploration is used as regularization. Therefore, we introduced
a more independent mechanism, a discount in the reward during backpropagation. This reduces the absolute values of the rewards propagated
from longer snippets as they approach the root node compared with those of shorter snippets that obtained the same reward.

\subsection{4.2 Using a Monte-Carlo Tree Search variant}

Our implementation \cite{basaldua2020arc} uses an MCTS algorithm \cite{browne2012survey} using the canonical UCB function as described
in equation 1 with one major modification. The simulation stage is replaced by a prediction of the reward using the reward function.

Algorithms for which an accurate simulator of the environment is not available, as it is the case of most real-world problems, cannot
implement the simulation phase and have to replace it by the evaluation of some model. The field is known as
\emph{model-based reinforcement learning} a notable example is \cite{schrittwieser2019mastering}.

\begin{figure}[h]
    \centering
    \includegraphics[width=0.45\textwidth]{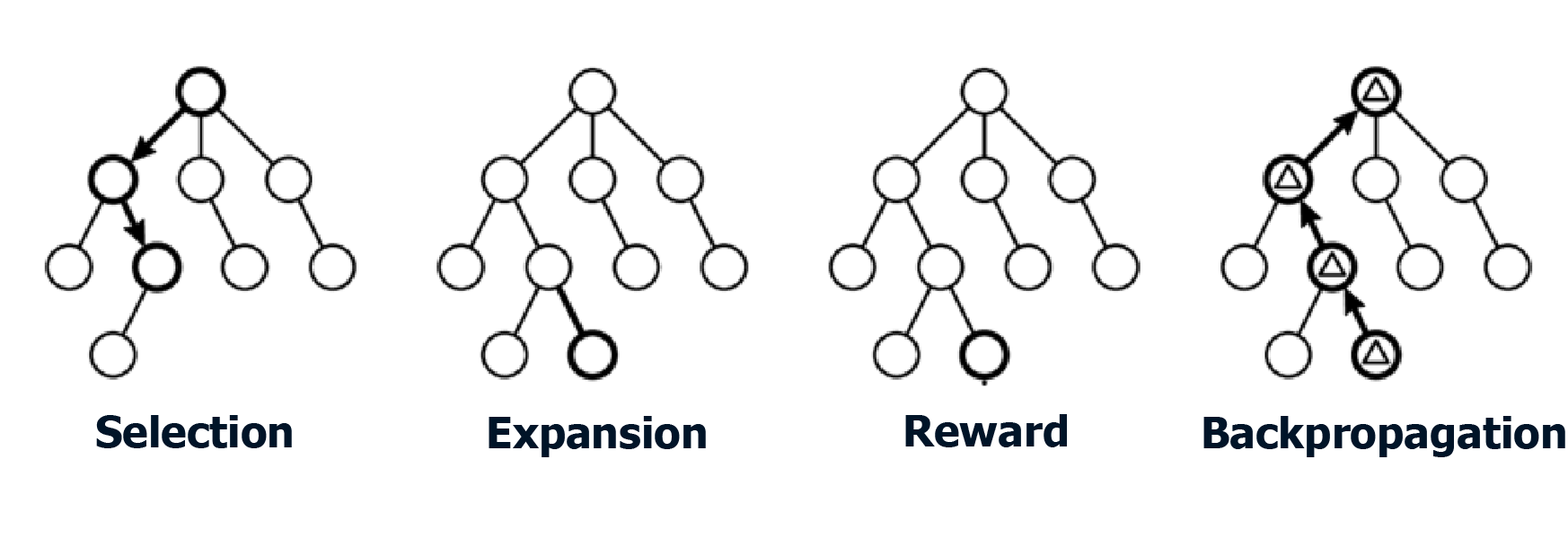}
    \caption{MCTS with a computed reward}
    \label{fig:mcts}
\end{figure}

Figure \ref{fig:mcts} shows the phases of the model-based MCTS algorithm. Most details have been discussed in sections 3 and 4 and the
open source implementation can be checked for precise implementation details.

\subsection{4.3 Other ideas for search}

\subsubsection{4.3.1 MuZero}

MuZero \cite{schrittwieser2019mastering} has been very influential in the design of the search. In many ways we opted for minimalism
and wanted to gather experience using "lower hanging" evaluation and reward functions, but we plan to implement more deep learning
into the framework, including a MuZero engine.

\subsubsection{4.3.2 Learning with self-modifying policies}

In model-based RL, we do not have a simulation policy anymore, but we do have a tree policy. At the moment, both priors and constants in
code snippets are static, re-using knowledge from the tree search to fit these constants has a huge potential. The idea of RL with
self-modifying policies is described in \cite{schmidhuber1998reinforcement} as "A learner's modifiable components are called its policy.
An algorithm that modifies the policy is a learning algorithm. If the learning algorithm has modifiable components represented as part of
the policy, then we speak of a self-modifying policy (SMP)". These ideas have already been applied to MCTS in go \cite{basaldua2013two}.

\subsubsection{4.3.3 Training an autoencoder}

The value function in our implementation is handcrafted, that requires domain knowledge and is probably worse than a deep learned function.
Exploiting the fact that we have snippet results and ground truth values in the same kind, we could train an autoencoder to learn how to
transform the former into the latter. The intermediate k-dimensional embedding becomes the value space. Then, we take only the encoding part
and use it as a value function.

\section{5. Experimental results}

We very briefly describe the challenge our open source Python implementation tackles. Detailed results including code snippets, running
times, solutions, parameters, etc. can be found in \cite{basaldua2020arc}.

\subsection{5.1 The ARC challenge}

\cite{chollet2019measure} approaches intelligence as "skill-acquisition efficiency". Rather than measuring it using performance at specific
tasks, he advocates for performance at tasks never seen before.

To test state of the art algorithms (or humans) in general problem solving, \cite{chollet2019arc} created a set of intelligence tests in
which an agent is expected to learn by a few examples a pattern that explains how some pictures were transformed into a given solution.
Then, a new picture is shown without the solution. The agent is expected to apply the learned pattern to produce the hidden solution. Only
a 100\% perfect match is considered valid.

This challenge was proposed as a three month long Kaggle competition finishing May 28, 2020, where solutions, submitted as code, ran over
100 unseen problems.

Since during three months almost a thousand participants were allowed to submit solutions three times a day, by the end of the competition
the tasks were not really unseen. Except for the winner, who built a search to create four instruction long programs in a domain specific
language of his invention, all other submitted solutions were ensembles of "whatever worked" favored by the possibility of team merging of
solutions found by trial. This explanation is based on what the authors themselves shared on the Kaggle platform. The winner mentioned
spending time optimizing multi-core C++ code for the search part and guessing what the unseen ideas could be to implement them in his
language. His code solved 20 out of the 100 hidden problems.

\subsection{5.2 Results}

We took a completely different approach given the short time frame and the need to implement everything. We did not give too much thought
to the FSL. On the public dataset of 800 problems we built something that could solve a bunch of easy problems and then, gradually,
extended it to some less easy cases. We finished our implementation four days before the deadline and ran 5 experiments from the May 24
to 27.

\begin{table}[h!]
\centering
\begin{tabular}{|c c|}
\hline
\multicolumn{2}{|c|}{Result summary} \\
\hline
Total running time (hours) & 70 \\
Problems believed to be solved but "wrong" & 4 \\
Total problems solved & 168 \\
Total previously unsolved problems solved & 56 \\
Unique previously unsolved problems solved & 22 \\
\hline
\end{tabular}
\caption{Aggregated results from five experiments using different settings.}
\label{table:results}
\end{table}

The system performance was very slow, a single threaded pure Python implementation, somehow mitigated by 70 hours of total runtime. Some of
the snippets labeled as "wrong" actually helped fixing the dataset. The experiments focused mostly on unsolved problems with a few already
solved problems as controls. The search solved controls in different ways than the snippets already in the codebase. The total number of
previously unsolved problems found, not counting repetitions across experiments, was 22. We would have considered anything above zero a
success. Table \ref{table:results} shows aggregated results.

There is no way to make a fair comparison with what other participants achieved. It would be necessary to know the size of the set of
solvable problems for our unrefined and incomplete FSL. We know it can solve at least 131 of 800 problems, because that is the size
of the codebase including the snippets found in the experiments. It is hard to code new solutions without extending the FSL, 150 could
be a reasonable estimate. All other problems are just impossible without extending the FSL. To be fair, the same would happen to
participants who built their own domain specific languages.

\section{6. Discussion}

So far, we have completed the description and analysis of the formal fields framework. We finish with some thoughts on how the framework
fits into our idea for multi-domain lifelong reasoning systems.

\subsection{6.1 Towards a pragmatic definition of understanding}

Informally, before even defining what \emph{understanding} is, most people, especially from outside the field, would doubt computers can
understand at all. We could give them examples from a narrow field, classic computer chess, and give them the opportunity to experience
how a simple minimax search, as it gets deeper, shows intelligent behavior. The program discovers strategy that has not been taught to
it, easily beats its author, etc. The chess engine can be used to play, but also to solve problems, create new problems, return insights
on strategic questions. One could write a five hundred page chess treaty from the discoveries made by a program with fifty lines of code
defining the rules of the game. The same people would probably concede, ”Okay, the computer understands chess, but that is a narrow domain,
it is not self-aware and would not invent a chess metaphor applied to another field.”. We agree that is a fair point, just let's walk one
step at a time and discuss what "the computer understands chess" means.

As a first approach:

\begin{enumerate}
  \item We have a formal field named \emph{chess} in which a language to write functions, a game state and actions have definitions.
  \item We have a function $f(s) \to A$ named \emph{the rules} returning a set of actions $a \in A$ (legal moves, priors, \dots) for
each game state $s$.
  \item We have an algorithm named \emph{minimax-search}.
\end{enumerate}

We could say: "\emph{Minimax-search} understands \emph{the rules} within the field of \emph{chess}." Why is minimax-search so smart it
can understand? Because it is an optimization algorithm, it can run, evaluate, optimize, fit, \dots. It is \emph{understanding} the
function since it is applying it and considers all implications relevant to optimizing. It is easy to argue this is a very narrow
definition of understanding, but it is hard to argue this is not understanding. A very narrow definition is all we need. By abstracting
the problem away, we have converted it into an optimization problem, something a computer definitely \emph{shows evidence of understanding}.

Why do we need such a thing at all? Is it just an intellectual tour de force with no practical consequences?

\begin{enumerate}
  \item In the first place, because we can. We already have a pragmatic definition of \emph{learning} (see 1.2) and most people would agree
it  captures  the essence of what learning is. Machine learning is a mainstream concept nowadays, it is not controversial that machines can
learn. We had "bad experiences" with semantics in the past and not everyone would agree that a computer can understand a symbol, but a
function in an optimization algorithm, that is something computers do understand. We can only make progress in \emph{machine understanding}
if we build a definition and then improve it if necessary.
  \item More importantly, because we need it. The opposite of understanding is "not understanding" and that is not the same as "not
knowing". That distinction is key to advance AI. Humans also don't understand many things and learn to live with it from the cradle. As we
advance representing ideas as code, we need a way to establish this difference. Maybe, while we still represent word embeddings as vectors,
we don't need it. Could we say "I don't understand this vector."? But then, we only have linear algebra to represent world semantics. If we
represent words as code, we have unlimited semantic representation and we can expect a machine to naturally say such things as: "I don't understand
this sentence, it doesn't run, I don't know that word, "optimistic molecule" doesn't make sense, \dots"
\end{enumerate}

Summarizing, it is always hard to coin an existing word with many nuances into a new formal definition. We argue that it is useful and
necessary. It clearly outlines the difference between not knowing and not understanding in a similar way humans do. It is also similar
to the accepted definition of understanding used in psychology: "being able to \emph{apply} knowledge rather than just recalling it"
\cite{bloom1956taxonomy}.

\subsection{6.2 Towards building lifelong learning systems}

Formal fields are not intended to be seen as just an algorithm that solves problems in many domains. It is a building block to create
lifelong learning systems that integrate all the domains at the same time, using one language grammar, one search algorithm and
different FSLs.

The idea has been proven to work and learn from as few examples as a human would need. The Jazz implementation \cite{basaldua2017jazz}
under development will provide at least a two decimal orders of magnitude speed improvement, smaller footprint, efficient storage to save,
fork or resume the state of any process and scalability across a cluster using technology already implemented in the platform.

Despite having been used on the abstract reasoning challenge first, most ideas have been thought of and tested with natural language in
mind since the first open source release of the Jazz platform, Dec 2017. Natural language itself can be seen as code, text as a sequence
of words and punctuation that are executable in some field assuming they can be \emph{understood within the field}. The combination of
handcrafted codebase snippets serving as examples, being extended gradually by ACG and the creation of more fields and concepts without
having to discard previous knowledge, but building on top of it, is the vision. To reach it, we will have to do more research and software
development.

\section{7. Conclusion}

In the first four sections we have introduced a narrow definition of ACG, presented the formal fields framework, discussed the language
design and analyzed the efficiency of the search algorithm. The experimental results applied on one of the hardest known challenges,
abstract reasoning, show that the whole idea works. All the experimental section is reproducible, code and data, and open sourced
in \cite{basaldua2020arc}.

Besides this proof of concept implementation, we have discussed future lines of research to build complex multi-domain lifelong learning
systems on top of the software stack included in the Jazz platform \cite{basaldua2017jazz}.

We also advocate for introducing a pragmatic definition of understanding (see 6.1) as a necessary notion in AI systems in which concepts
are represented by code.

\section{8. Acknowledgments}
We wish to thank BBVA for releasing Jazz version 0.1.7 as open source software. Jazz is a highly efficient data processing platform
that is currently being refactored as a server implementing formal fields in both research and industrialized applications.

\bibliographystyle{aaai}
\bibliography{ref}

\begin{thebibliography}{}

\bibitem[\protect\citeauthoryear{Abelson, Sussman, and
  Sussman}{1996}]{abelson1996structure}
Abelson, H.; Sussman, G.~J.; and Sussman, J.
\newblock 1996.
\newblock {\em Structure and interpretation of computer programs}.
\newblock Justin Kelly.

\bibitem[\protect\citeauthoryear{Baader \bgroup et al\mbox.\egroup
  }{2003}]{baader2003description}
Baader, F.; Calvanese, D.; McGuinness, D.; Patel-Schneider, P.; Nardi, D.;
  et~al.
\newblock 2003.
\newblock {\em The description logic handbook: Theory, implementation and
  applications}.
\newblock Cambridge university press.

\bibitem[\protect\citeauthoryear{Balog \bgroup et al\mbox.\egroup
  }{2016}]{balog2016deepcoder}
Balog, M.; Gaunt, A.~L.; Brockschmidt, M.; Nowozin, S.; and Tarlow, D.
\newblock 2016.
\newblock Deepcoder: Learning to write programs.
\newblock {\em arXiv preprint arXiv:1611.01989}.

\bibitem[\protect\citeauthoryear{Basald{\'u}a \bgroup et al\mbox.\egroup
  }{2013}]{basaldua2013two}
Basald{\'u}a, J.; Stewart, S.; Moreno-Vega, J.~M.; and Drake, P.~D.
\newblock 2013.
\newblock Two online learning playout policies in monte carlo go: An
  application of win/loss states.
\newblock {\em IEEE Transactions on Computational Intelligence and AI in Games}
  6(1):46--54.

\bibitem[\protect\citeauthoryear{Basald{\'u}a}{2017}]{basaldua2017jazz}
Basald{\'u}a, J.
\newblock 2017.
\newblock Jazz, the efficient open source ai platform.
\newblock \url{https://github.com/kaalam/Jazz}.

\bibitem[\protect\citeauthoryear{Basald{\'u}a}{2020}]{basaldua2020arc}
Basald{\'u}a, J.
\newblock 2020.
\newblock Jazzarc, poc on code generation using formal fields on the arc
  challenge.
\newblock \url{https://github.com/kaalam/JazzARC}.

\bibitem[\protect\citeauthoryear{Bellman}{1957a}]{bellman1957dynamic}
Bellman, R.
\newblock 1957a.
\newblock Dynamic programming.
\newblock {\em Princeton UniversityPress. BellmanDynamic programming1957}  151.

\bibitem[\protect\citeauthoryear{Bellman}{1957b}]{bellman1957markovian}
Bellman, R.
\newblock 1957b.
\newblock A markovian decision process.
\newblock {\em Journal of mathematics and mechanics}  679--684.

\bibitem[\protect\citeauthoryear{Bloom and others}{1956}]{bloom1956taxonomy}
Bloom, B.~S., et~al.
\newblock 1956.
\newblock Taxonomy of educational objectives. vol. 1: Cognitive domain.
\newblock {\em New York: McKay} 20:24.

\bibitem[\protect\citeauthoryear{Bottou}{2014}]{bottou2014machine}
Bottou, L.
\newblock 2014.
\newblock From machine learning to machine reasoning.
\newblock {\em Machine learning} 94(2):133--149.

\bibitem[\protect\citeauthoryear{Browne \bgroup et al\mbox.\egroup
  }{2012}]{browne2012survey}
Browne, C.~B.; Powley, E.; Whitehouse, D.; Lucas, S.~M.; Cowling, P.~I.;
  Rohlfshagen, P.; Tavener, S.; Perez, D.; Samothrakis, S.; and Colton, S.
\newblock 2012.
\newblock A survey of monte carlo tree search methods.
\newblock {\em IEEE Transactions on Computational Intelligence and AI in games}
  4(1):1--43.

\bibitem[\protect\citeauthoryear{Chen and Guestrin}{2016}]{chen2016xgboost}
Chen, T., and Guestrin, C.
\newblock 2016.
\newblock Xgboost: A scalable tree boosting system.
\newblock In {\em Proceedings of the 22nd acm sigkdd international conference
  on knowledge discovery and data mining},  785--794.

\bibitem[\protect\citeauthoryear{Chollet}{2019a}]{chollet2019arc}
Chollet, F.
\newblock 2019a.
\newblock The abstraction and reasoning corpus (arc).
\newblock \url{https://github.com/fchollet/ARC}.

\bibitem[\protect\citeauthoryear{Chollet}{2019b}]{chollet2019measure}
Chollet, F.
\newblock 2019b.
\newblock The measure of intelligence.
\newblock {\em arXiv preprint arXiv:1911.01547}.

\bibitem[\protect\citeauthoryear{Church}{1936}]{church1936unsolvable}
Church, A.
\newblock 1936.
\newblock An unsolvable problem of elementary number theory.
\newblock {\em American journal of mathematics} 58(2):345--363.

\bibitem[\protect\citeauthoryear{d'Autume \bgroup et al\mbox.\egroup
  }{2019}]{d2019episodic}
d'Autume, C. d.~M.; Ruder, S.; Kong, L.; and Yogatama, D.
\newblock 2019.
\newblock Episodic memory in lifelong language learning.
\newblock {\em arXiv preprint arXiv:1906.01076}.

\bibitem[\protect\citeauthoryear{Devlin \bgroup et al\mbox.\egroup
  }{2017}]{devlin2017robustfill}
Devlin, J.; Uesato, J.; Bhupatiraju, S.; Singh, R.; Mohamed, A.-r.; and Kohli,
  P.
\newblock 2017.
\newblock Robustfill: Neural program learning under noisy i/o.
\newblock In {\em Proceedings of the 34th International Conference on Machine
  Learning-Volume 70},  990--998.
\newblock JMLR. org.

\bibitem[\protect\citeauthoryear{Garnelo, Arulkumaran, and
  Shanahan}{2016}]{garnelo2016towards}
Garnelo, M.; Arulkumaran, K.; and Shanahan, M.
\newblock 2016.
\newblock Towards deep symbolic reinforcement learning.
\newblock {\em arXiv preprint arXiv:1609.05518}.

\bibitem[\protect\citeauthoryear{Gaunt \bgroup et al\mbox.\egroup
  }{2016}]{gaunt2016terpret}
Gaunt, A.~L.; Brockschmidt, M.; Singh, R.; Kushman, N.; Kohli, P.; Taylor, J.;
  and Tarlow, D.
\newblock 2016.
\newblock Terpret: A probabilistic programming language for program induction.
\newblock {\em arXiv preprint arXiv:1608.04428}.

\bibitem[\protect\citeauthoryear{Graves \bgroup et al\mbox.\egroup
  }{2016}]{graves2016hybrid}
Graves, A.; Wayne, G.; Reynolds, M.; Harley, T.; Danihelka, I.;
  Grabska-Barwi{\'n}ska, A.; Colmenarejo, S.~G.; Grefenstette, E.; Ramalho, T.;
  Agapiou, J.; et~al.
\newblock 2016.
\newblock Hybrid computing using a neural network with dynamic external memory.
\newblock {\em Nature} 538(7626):471--476.

\bibitem[\protect\citeauthoryear{Graves, Wayne, and
  Danihelka}{2014}]{graves2014neural}
Graves, A.; Wayne, G.; and Danihelka, I.
\newblock 2014.
\newblock Neural turing machines.
\newblock {\em arXiv preprint arXiv:1410.5401}.

\bibitem[\protect\citeauthoryear{Harnad}{1990}]{harnad1990symbol}
Harnad, S.
\newblock 1990.
\newblock The symbol grounding problem.
\newblock {\em Physica D: Nonlinear Phenomena} 42(1-3):335--346.

\bibitem[\protect\citeauthoryear{Heule \bgroup et al\mbox.\egroup
  }{2016}]{heule2016stratified}
Heule, S.; Schkufza, E.; Sharma, R.; and Aiken, A.
\newblock 2016.
\newblock Stratified synthesis: automatically learning the x86-64 instruction
  set.
\newblock In {\em Proceedings of the 37th ACM SIGPLAN Conference on Programming
  Language Design and Implementation},  237--250.

\bibitem[\protect\citeauthoryear{Hudson and Manning}{2019}]{hudson2019learning}
Hudson, D.~A., and Manning, C.~D.
\newblock 2019.
\newblock Learning by abstraction: The neural state machine.
\newblock {\em arXiv preprint arXiv:1907.03950}.

\bibitem[\protect\citeauthoryear{Kaiser \bgroup et al\mbox.\egroup
  }{2017}]{kaiser2017one}
Kaiser, L.; Gomez, A.~N.; Shazeer, N.; Vaswani, A.; Parmar, N.; Jones, L.; and
  Uszkoreit, J.
\newblock 2017.
\newblock One model to learn them all.
\newblock {\em arXiv preprint arXiv:1706.05137}.

\bibitem[\protect\citeauthoryear{Kant}{2018}]{kant2018recent}
Kant, N.
\newblock 2018.
\newblock Recent advances in neural program synthesis.
\newblock {\em arXiv preprint arXiv:1802.02353}.

\bibitem[\protect\citeauthoryear{Kocsis and
  Szepesv{\'a}ri}{2006}]{kocsis2006bandit}
Kocsis, L., and Szepesv{\'a}ri, C.
\newblock 2006.
\newblock Bandit based monte-carlo planning.
\newblock In {\em European conference on machine learning},  282--293.
\newblock Springer.

\bibitem[\protect\citeauthoryear{Lachaux \bgroup et al\mbox.\egroup
  }{2020}]{lachaux2020unsupervised}
Lachaux, M.-A.; Roziere, B.; Chanussot, L.; and Lample, G.
\newblock 2020.
\newblock Unsupervised translation of programming languages.
\newblock {\em arXiv preprint arXiv:2006.03511v1}.

\bibitem[\protect\citeauthoryear{Manna and Waldinger}{1971}]{manna1971toward}
Manna, Z., and Waldinger, R.~J.
\newblock 1971.
\newblock Toward automatic program synthesis.
\newblock {\em Communications of the ACM} 14(3):151--165.

\bibitem[\protect\citeauthoryear{Marcus}{2018}]{marcus2018deep}
Marcus, G.
\newblock 2018.
\newblock Deep learning: A critical appraisal.
\newblock {\em arXiv preprint arXiv:1801.00631}.

\bibitem[\protect\citeauthoryear{McCarthy}{1978}]{mccarthy1978history}
McCarthy, J.
\newblock 1978.
\newblock History of lisp.
\newblock In {\em History of programming languages}. Association for Computing
  Machinery.
\newblock  173--185.

\bibitem[\protect\citeauthoryear{McCloskey and
  Cohen}{1989}]{mccloskey1989catastrophic}
McCloskey, M., and Cohen, N.~J.
\newblock 1989.
\newblock Catastrophic interference in connectionist networks: The sequential
  learning problem.
\newblock In {\em Psychology of learning and motivation}, volume~24. Elsevier.
\newblock  109--165.

\bibitem[\protect\citeauthoryear{Miller}{1998}]{miller1998introduction}
Miller, E.
\newblock 1998.
\newblock An introduction to the resource description framework.
\newblock {\em Bulletin of the American Society for Information Science and
  Technology} 25(1):15--19.

\bibitem[\protect\citeauthoryear{Minsky}{1967}]{minsky1967programming}
Minsky, M.
\newblock 1967.
\newblock Why programming is a good medium for expressing poorly understood and
  sloppily formulated ideas.
\newblock {\em Design and planning II-computers in design and communication}
  120--125.

\bibitem[\protect\citeauthoryear{Minsky}{1968}]{minsky1968semantic}
Minsky, M.~L.
\newblock 1968.
\newblock {\em Semantic information processing}.
\newblock MIT Press.

\bibitem[\protect\citeauthoryear{Pu \bgroup et al\mbox.\egroup
  }{2017}]{pu2017selecting}
Pu, Y.; Miranda, Z.; Solar-Lezama, A.; and Kaelbling, L.~P.
\newblock 2017.
\newblock Selecting representative examples for program synthesis.
\newblock {\em arXiv preprint arXiv:1711.03243}.

\bibitem[\protect\citeauthoryear{Rakotoarison, Schoenauer, and
  Sebag}{2019}]{rakotoarison2019automated}
Rakotoarison, H.; Schoenauer, M.; and Sebag, M.
\newblock 2019.
\newblock Automated machine learning with monte-carlo tree search (extended
  version).
\newblock {\em arXiv preprint arXiv:1906.00170}.

\bibitem[\protect\citeauthoryear{Real \bgroup et al\mbox.\egroup
  }{2020}]{real2020automl}
Real, E.; Liang, C.; So, D.~R.; and Le, Q.~V.
\newblock 2020.
\newblock Automl-zero: Evolving machine learning algorithms from scratch.
\newblock {\em arXiv preprint arXiv:2003.03384}.

\bibitem[\protect\citeauthoryear{Schmidhuber, Zhao, and
  Schraudolph}{1998}]{schmidhuber1998reinforcement}
Schmidhuber, J.; Zhao, J.; and Schraudolph, N.~N.
\newblock 1998.
\newblock 12. reinforcement learning with self-modifying policies.
\newblock In {\em Learning to learn}. Springer.
\newblock  293--309.

\bibitem[\protect\citeauthoryear{Schrittwieser \bgroup et al\mbox.\egroup
  }{2019}]{schrittwieser2019mastering}
Schrittwieser, J.; Antonoglou, I.; Hubert, T.; Simonyan, K.; Sifre, L.;
  Schmitt, S.; Guez, A.; Lockhart, E.; Hassabis, D.; Graepel, T.; et~al.
\newblock 2019.
\newblock Mastering atari, go, chess and shogi by planning with a learned
  model.
\newblock {\em arXiv preprint arXiv:1911.08265}.

\bibitem[\protect\citeauthoryear{Silver, Yang, and
  Li}{2013}]{silver2013lifelong}
Silver, D.~L.; Yang, Q.; and Li, L.
\newblock 2013.
\newblock Lifelong machine learning systems: Beyond learning algorithms.
\newblock In {\em 2013 AAAI spring symposium series}.

\bibitem[\protect\citeauthoryear{Sowa}{1987}]{sowa1987semantic}
Sowa, J.~F.
\newblock 1987.
\newblock Semantic networks.
\newblock {\em Encyclopedia of Artificial Intelligence, second edition, 1992}.

\bibitem[\protect\citeauthoryear{Sutton and
  Barto}{1998}]{sutton1998reinforcement}
Sutton, R.~S., and Barto, A.~G.
\newblock 1998.
\newblock {\em Reinforcement learning: An introduction}, volume~1.
\newblock MIT press Cambridge.

\bibitem[\protect\citeauthoryear{Taddeo and Floridi}{2005}]{taddeo2005solving}
Taddeo, M., and Floridi, L.
\newblock 2005.
\newblock Solving the symbol grounding problem, a critical review of fifteen
  years of research.
\newblock {\em Journal of Experimental \& Theoretical Artificial Intelligence}
  17(4):419--445.

\bibitem[\protect\citeauthoryear{Thrun}{1998}]{thrun1998lifelong}
Thrun, S.
\newblock 1998.
\newblock Lifelong learning algorithms.
\newblock In {\em Learning to learn}. Springer.
\newblock  181--209.

\bibitem[\protect\citeauthoryear{Turing}{1937}]{turing1937computable}
Turing, A.~M.
\newblock 1937.
\newblock {\em On computable numbers, with an application to the
  Entscheidungsproblem; a correction}.
\newblock Royal Society.

\bibitem[\protect\citeauthoryear{Zaremba and
  Sutskever}{2015}]{zaremba2015reinforcement}
Zaremba, W., and Sutskever, I.
\newblock 2015.
\newblock (revised)reinforcement learning neural turing machines.
\newblock {\em arXiv preprint arXiv:1505.00521}.

\bibitem[\protect\citeauthoryear{Zaremba \bgroup et al\mbox.\egroup
  }{2016}]{zaremba2016learning}
Zaremba, W.; Mikolov, T.; Joulin, A.; and Fergus, R.
\newblock 2016.
\newblock Learning simple algorithms from examples.
\newblock In {\em International Conference on Machine Learning},  421--429.

\end{thebibliography}

\end{document}